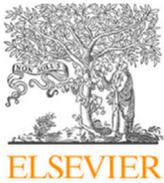
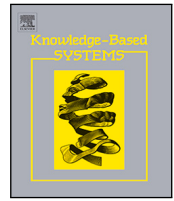

# Evaluation of semantic relations impact in query expansion-based retrieval systems

Lorenzo Massai

*Università degli Studi di Firenze, Department of Mathematics and Computer Science, Firenze, Italy*

A B S T R A C T

With the increasing demand of intelligent systems capable of operating in different contexts (e.g. users on the move) the correct interpretation of the user-need by such systems has become crucial to give consistent answers to the user questions. The most effective applications addressing such task are in the fields of natural language processing and semantic expansion of terms. These techniques are aimed at estimating the goal of an input query reformulating it as an intent, commonly relying on textual resources built exploiting different semantic relations like *synonymy*, *antonymy* and many others. The aim of this paper is to generate such resources using the labels of a given taxonomy as source of information. The obtained resources are integrated into a plain classifier for reformulating a set of input queries as intents and tracking the effect of each relation, in order to quantify the impact of each semantic relation on the classification. As an extension to this, the best tradeoff between improvement and noise introduction when combining such relations is evaluated. The assessment is made generating the resources and their combinations and using them for tuning the classifier which is used to reformulate the user questions as labels. The evaluation employs a wide and varied taxonomy as a use-case, exploiting its labels as basis for the semantic expansion and producing several corpora with the purpose of enhancing the pseudo-queries estimation.

## 1. Introduction

Information retrieval architectures often deal with the interpretation of the user query, especially when it comes to queries which can be posed in natural language. The main hold back is the identification of a connection between unstructured questions and structured data, which is an open problem and main task of natural language processing applications. The estimation of the user intent in natural language queries can be a difficult task and it cannot be achieved considering only the syntactic representation of terms. A semantic characterization is needed to reformulate an unconstrained question as part of a structure, employing the sense and the context of terms. The applications to get such characterizations are mainly in the fields of machine learning and semantic expansion, addressing problems like context learning, term disambiguation and polysemy. Semantic characterization of terms can be achieved, among the others, through data aggregation exploiting automatic clustering techniques or ontologies, which provide a qualitative representation of documents through semantic data models. The different strategies are analyzed in Section 2. The aim of this paper is to propose, tune and evaluate an automatic procedure to generate semantic resources which can be exploited by query expansion-based retrieval systems, showing the most effective combination of the generated resources. The focus of the research is on the evaluation, demonstrating the implications of employing different semantic expansions when associating a natural language query to a pre-defined class. There are four main contributions of this work:

- We propose a flexible approach to reformulate user questions into classes of an arbitrary taxonomy, employing a modular architecture which allows to evaluate different strategies of building the semantic resources upon which the architecture is based.
- An automatic procedure for generating the resources based on different semantic relations is described and developed.
- An automatic procedure to combine the different resources which have been generated is described and developed, aiming at identifying the inter-relations among them.
- We provide an estimation of the impact of each combination for the considered task, identifying the most effective semantic relation and the most effective combination of them.

All the data exploited for the assessment and the code produced for generating the resources and building the classification pipeline are made available within the paper for replicating the experiments. The procedure for generating the semantic resources which are employed in the experiments is based on state-of-the-art semantic repositories and is described in Section 3, as well as the different strategies for building them. In Section 4 a modular classification system is built and evaluated exploiting the different sets of knowledge resources obtained in Section 3, showing both the most impacting relation and the most effective combination of them. The evaluation is assessed loading the generated corpora into a GATE pipeline to measure the best and worst performance metrics among all the configurations. The data resources which are used to load the classifiers for the evaluation are chosen in the range of Points of Interest (POIs) search engines.

## 2. Related work

In the field of query reformulation and features augmentation a key role is held by semantic networks and semantic repositories. The relations among terms can be used to get the context and sense behind a word in a phrase and the analysis of word frequencies in wide knowledge bases can be exploited as well to get latent semantics. The target of query expansion is giving relevance not only to the syntactic form of the words which are present within the query, but also to the most representative meaning of the words expressing the user-need. The main approaches for term augmentation can be divided in statistical [1], semantic [2] and hybrid [3] and the main techniques are interactive query expansion, pseudo-relevance feedback, search results clustering and semantic networks. In this section the most widespread approaches to term expansion are reviewed.

### 2.1. Query expansion techniques

Query expansion is a field of natural language processing which aim is to provide an augmentation of the input terms composed by their most related concepts. A query expansion tool can be exploited to improve the retrieval capabilities of a search engine. There are many approaches to get such augmentation which can be classified as global analysis and local analysis [4]. The global methods are independent from the query and include interactive and automatic expansions. Local methods use the original query to retrieve documents and the query terms are augmented by the means of a relevance feedback made by users or estimated on the result-set. Semantic networks can be exploited to get closer to the semantics of the query terms considering the sense of each term and its classification [5] and also to obtain expansion terms from concept relations [6]. Dey et al. [7] show a 41,5% and 22% increase in precision expanding search queries with the vocabularies of plant and wine ontologies, choosing the terms with the shortest semantic distance between query terms and ontology terms. A similar effect can be achieved through search result clustering, which can be performed to estimate the classes (concepts) considering the semantic relatedness between terms as distances between them. Also, latent semantic analysis techniques [8] can be used to expand the feature space estimating the concepts which are related to terms by the relations among them.

**Interactive query expansion** is a query expansion technique which consists in expanding the original query and getting a feedback from the user on the relevance of a retrieved set of documents [9]. The set of retrieved documents is evaluated by the user expressing a choice on the best match for his/her need and explicitly refining the query formulation (Fig. 1). Such method is effective to meet the user-need, but it is depending on user interaction and context which are, in general, prone to ambiguity.

**Pseudo-relevance feedback** (PRF) is a refinement of relevance feedback which adds an automated step to avoid the user contribution in the process of query expansion [10]. The data flow from the user query to the retrieved documents includes the variation of the weights of query terms assuming that the top retrieved documents are relevant and using the position of the documents as a feedback for relevance (this makes automatic the step *c.* in Fig. 1). Pseudo-relevance feedback relies on the top documents retrieved by the logic itself, hence the method is heavily dependent on the ranking algorithm; for this reason [11] in their pseudo-relevance approach explore different weighting algorithms to weigh the terms.

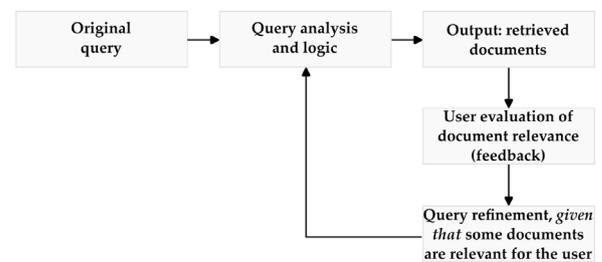

**Fig. 1.** Relevance feedback.

A classic application of relevance feedback is the Rocchio algorithm, a nearest centroid technique for classification that assigns to documents the class label of the training samples whose mean is closest to the document's class, modifying the weights of the vectors representing both the query and the documents. The performance of the algorithm is noticeable respect to other expansion techniques, as demonstrated in [12] over each TREC track . Rocchio classification can sometimes have an unwanted behavior reflecting the importance of a document respect to the whole collection rather than to the user query; some studies have mitigated this drawback including in the relevance score also documents with high ranking outside of the top relevant documents [13]. The adaptability of pseudo-relevance feedback mechanisms to recent transformer-based technologies like BERT [14] has been studied in [15], where a representative feedback is extracted from the embeddings showing promising performances respect to traditional approaches. Wang et al. [16] developed a pseudo-relevance feedback framework exploiting probability-based PRF methods and language-model-based PRF methods, combining relevance and semantic matching to re-rank the semantic similarity between the query and document representations.

**Search result clustering** consists in partitioning the set of documents to retrieve in similarity classes and using them as an index for the actual retrieval. Text clustering aims at grouping the documents to retrieve in semantically related sets and it is usually performed online. Such task can be addressed using data-centric approaches or description-centric approaches. Data-centric approaches include data agglomeration techniques like K-means and its variants [17] and SVM classifiers which have been demonstrated suitable for both binary and multi-class classification problems [18]. Description-centric approaches are less dependent from data and include semantic agglomeration like ontology models and automatic taxonomy clustering [19]. The aggregation obtained through the clustering process can be considered as additional information expressing the semantics of the documents which are in the same cluster, therefore can be used as a feature for the original query terms. Such strategy is presented in [20], in which the cluster labels are used as an expansion of the original query terms. It is to be noticed that the expansion of query terms can potentially result into a deviance from the user-need known as query drift [21] since automatic expansion deals with polysemy, term disambiguation and user context [22,23] when determining the most representative meaning of terms. Cao et al. [24] evaluate the usefulness of expansion terms, highlighting that only a small portion of the traditional expansion terms have impact on the retrieval effectiveness. To limit query drift, a tradeoff between improvement and worsening of a retrieval system relying on query expansions has to be evaluated.

### 2.2. Semantic networks and knowledge bases

A **semantic network**, or **frame network** is an abstract representation of concepts and relations between them. The usual representation of a semantic network is a directed or undirected graph containing nodes as concepts and edges as relations (Fig. 2); such representation aims at expressing the semantics of terms in a domain. The conceptualization of a term is strictly related to the context in which it is used and which could not be expressed by the term itself. In semantic networks the context of an entity is expressed by its relations, which are essential to differentiate one over another. As an example, in a semantic network describing the biology context the term *tree* will occur having the relation of hypernymy linked to the term *oak*; in the context of algebra the same term *tree*, which expresses an *algebraic structure*, will not be associated to the term *oak*. The different semantic characterization expresses the difference between the two concepts and studying these relations can support semantic disambiguation of terms [25]. Recent efforts in deriving contextual information from domain ontologies to generate expansion terms can also be found in [26–28] which use domain-specific ontologies to implement query expansion mechanisms.

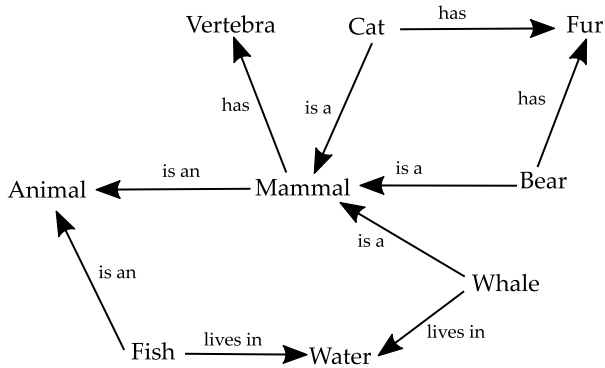

**Fig. 2.** Semantic network for concept *mammal*.

A **knowledge base** is an instantiation of data in a structured representation of concepts expressed through assertion statements (A-box) and terminology statements (T-box). A-box statements of an ontology represent factual assertions and relations among concrete entities (e.g. Rome *is-a* city), while T-box statements represent the properties of domain classes (e.g. city *contains* buildings) and make possible inference (e.g. Rome *contains* buildings). The data that are stored in a knowledge base can be retrieved by the properties defined in the T-box; this structure is also referred as a populated ontology. Query expansion typically relies both on semantic networks and knowledge bases to get the relations between concepts. Recent efforts in query augmentation based on ontologies and knowledge bases can be found in [27], combining the relevance score produced by different rank aggregation approaches and using fuzzy rules to infer the weights of the additional query terms.

*2.2.1. WordNet*

The most representative semantic network, though not the most recent, is the lexical database WordNet [29]. It groups English nouns, verbs, adjectives and adverbs into synsets, each expressing a distinct concept. Synsets are interlinked by the means of conceptual-semantic and lexical relations. WordNet is similar to a thesaurus, since it groups words together basing on their meanings. However, there are some important distinctions. First, WordNet interlinks not just the lexical form of words, but specific senses of words. As a result, words that are found to be near one to another in the network are semantically disambiguated. Second, WordNet labels the semantic relations among words, whereas the groupings of words in a thesaurus does not follow any explicit pattern other than meaning similarity. Some of the most representative semantic relations which are covered are *holonymy*, *hyponymy* (or *troponymy*), *synonymy*, *antonymy*, *hypernymy* and *meronymy*.

- **Synonymy** is the relation which includes the terms which are different from the input word, having the same meaning. The synonyms are obtained retrieving the terms contained within the same WordNet synset (e.g. computer, *desktop*).
- **Antonymy** is the relation which includes all the terms with opposite meaning with respect to the input word (e.g. black, *white*)
- **Hypernymy** is the relation which includes all the terms which constitute the more general category respect to the input word (e.g. dog, *animal*)
- **Hyponymy** is the relation which includes the terms with a similar, but less specific meaning (e.g. cd, *disc*)
- **Meronymy** is the relation which does not consider the input as its specific meaning, but of the more general concept it represents (e.g. head, *boss*)
- **Holonomy** is the opposite relation to meronymy and it not considers the input as its meaning, but the more specific concept it represents (e.g. boss, *head*)

Applications of query augmentation based on Wordnet can be found in [30–32]; an effective effort in extracting lexical information from WordNet relations can also be found in [33], combining automatic and user-supported expansions. The WordNet web endpoint is available at: https://wordnet.princeton.edu/ and can be used to access all the Wordnet semantic relations.

Recent works like Jain et al. [34] declare Wordnet *synonyms* the most effective semantic relation for query expansion and synonyms are likewise preferred in most query expansion tasks like Shi et al. [35] and Chauhan et al. [36], while their effectiveness respect to other semantic relations is never evaluated. Navigli and Velardi [37] show that expansions based on synonyms and hypernyms has limited effects on web information retrieval performance and that the most successful query expansion methods are based on words that co-occur with the query terms and word glosses. Works like Buscaldi et al. [38] use a combination of synonyms and meronyms, while others like Esposito et al. [39] and Franco-Salvador et al. [40] evaluate the employment of synonyms and hypernyms expansions. Several endpoints like Copeland et al. [41], available online through the Microsoft Azure Points Of Interest retrieval API, add to each category a small set of synonyms to improve the retrieval. Synonyms seem to be a constant choice in query expansion and also the most intuitive choice for related words identification, while their actual effectiveness respect to other expansions is not proven. The question has been also emerged in recent works like Azad and Deepak [42] and Pal et al. [43], where the problem of which semantic relation would be the most effective in query expansion tasks is expressed and the accent is posed on the use of synonyms and their effective enhancement when expanding the query terms.

*2.2.2. Babelnet*

The Babelnet [22] semantic network features query expansion and disambiguation through the generation of semantically related sets of words called synsets (cognitive synonyms). The Babelnet framework provides term augmentation and transformation of each query term into a synset gathering data from WordNet, Open Multilingual WordNet, OmegaWiki, Wiktionary, Wikidata and Wikipedia (Fig. 3). Each element of the synset is denoted as the entity which is most related to the concept of the term (i.e. its sense) within the Babelnet semantic graph. A concept is *most related* to another if it minimizes some distance (respect to each semantic relation) with any other contained in the networks outlined above.

*2.2.3. Google Books N-grams dictionaries*

Google Books is an endpoint for browsing the books data indexed by Google and it directly allows the retrieval of book texts. N-grams are fixed-size tuples of items, and in this case the items are words extracted from the Google Books corpus. The $N$ specifies the number of elements in the tuple: a 5-g contains five words or characters. The N-grams are produced by passing a sliding window on the text of books indexed by Google and giving as output a record for each new token. Such utility can be used for example to extract the terms that frequently surround an input word (e.g. *friend* → *best*). The Google Books N-grams data are indexed by several endpoints like the Datamuse API [44] (which also includes a Wordnet access point) and it is accessible configuring the API with the proper code. This endpoint can also be exploited to extract the popular nouns with input adjectives, the popular adjectives with input noun, the popular associated words, the frequent follower word and the frequent predecessor word.

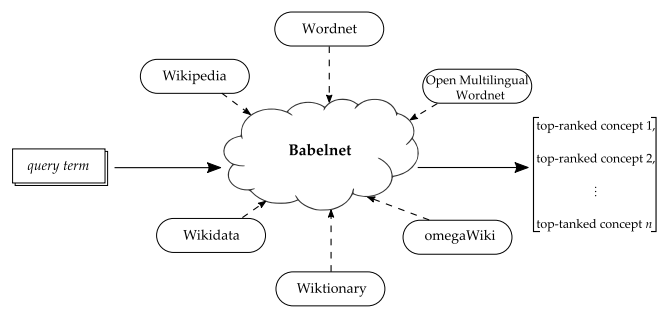

**Fig. 3.** Resources exploited by the Babelnet semantic expansion.

- The **popular nouns with input adjectives** is a relation between the input words, which are adjectives, and the nouns which the adjectives are referred to (e.g. brown *eyes*, hot *water*);
- The **popular adjectives with input noun** is the same as the former, with nouns as input (e.g. *brown* eyes, *hot* water);
- The **popular associated words** is a relation which links to the input terms the words often related to them (e.g. bottle of *water*, glass of *wine*);
- The **frequent follower word** is a relation which bridges the input terms to words which are often following them (good *morning*, welcome *to*);
- The **frequent predecessor word** is the same as the former, linking the input to the words which are often preceded by the input terms (*good* morning, *welcome* to).

*2.2.4. Onelook dictionaries*

The Onelook Dictionary [45] search is a web endpoint and API which allows the user to look for definitions and words related to the input words or phrase. The search engine indexes more than 19 million words in more than 1000 online dictionaries, like Oxford Dictionaries, Merriam-Webster, American Heritage, Collins English Dictionary, Macmillan Dictionary, Wiktionary, Online Etymology Dictionary, Wikipedia, Rhymezone, WordNet 1.7, Free Dictionary, Dictionary.com, Wordnik and many others. The dictionaries can be browsed by topic: art, business, computing, medicine, miscellaneous, religion, science, slang, sports, technology and the definitions can be restricted to Chinese, English, French, German, Italian and Spanish. Onelook dictionaries endpoint is available at: https://www.onelook.com/.

*2.3. NLP frameworks*

**Keras** is a high-level Python API based on neural networks that runs on CPU or GPU. It supports convolutional and recurrent networks and it can be used for text classification, text generation and summarization, tagging, parsing, machine translation, speech recognition, and other tasks. Keras runs on top of many frameworks and numerical computation libraries like **TensorFlow**, which is an open-source machine learning library by Google allowing GPU computation. The main concept behind TensorFlow is flow graphs usage: the nodes of the graph reflect mathematical operations, while the edges represent multidimensional inter-communicating data arrays (tensors). One of the most known TensorFlow's NLP applications is Google Translate. Other applications are text classification and summarization, speech recognition, tagging, and so on. Considering the neural networks approach, **PyTorch** is a fast library which builds neural networks on a tape-based autograd system and provides tensor computation with strong GPU acceleration. Recurrent neural networks are mostly used in PyTorch for machine translation, classification, text generation, tagging, and other NLP tasks. The **Dynet** neural network library by Carnegie Mellon University adds the feature of handling syntactic parsing, which makes it an attractive choice for deeper analysis of the phrase. It supports C++, Python languages and GPU computing. Dynet is based on the dynamic declaration of network structure and over syntactic parsing it addresses also machine translation and morphological inflection analysis.

Stanford's **Core NLP** is a flexible and fast grammatical analysis tool that provides APIs for most common programming languages including Python and Java; it can also be run as a web service. The framework has features like part-of-speech (POS) tagging, named entity recognizing (NER), parsing, co-reference resolution, sentiment analysis, bootstrapped pattern learning, and open information extraction tools. Looking at feature separation and modularity, **GATE** [46] is a Java framework for NLP which is organized as a pipeline: the elements of the pipeline can be added activating some features of the pipeline, which is run as a script. GATE provides lemmatization, POS-tagging, stemming, tokenization and many others, relying on low-level libraries which can be loaded in the pipeline like the multilingual TreeTagger POS-tagger [47], the Snowball stemmer [48] and the Stanford parser [49]. Looking at optimization and deep learning, **Deeplearning4j** is a Java programming library which can make use of the Keras API through TensorFlow. The main features of Deeplearning4j are the use of distributed CPUs and GPUs, parallel training via iterative reduce and micro-service architecture adaptation; moreover, vector space modeling enables solving of text-mining problems. It features POS-tagging, dependency parsing and word embedding through *word2vec*, a simple neural network which takes a corpus as input and computes vectors representing the semantic distribution of the words.

**3. Generation of the document sets**

The focus of this section is describing the procedures which have been employed to generate several sets of documents, each based on a specific semantic relation. The produced documents can be exploited to improve retrieval systems based on taxonomies, like ontology-based systems and frameworks driven by hierarchical structures. In particular, the GATE framework and the Datamuse endpoint have been employed in the resource generation phase for accessing semantic repositories (Sections 3.2 and 3.3) and also in the evaluation phase (Section 4) in which are used to classify a set of user queries into taxonomy classes.

*3.1. Technologies and resources*

To analyze the benefits and the drawbacks of employing a corpus built by the means of a specific semantic relation respect to the others, the GATE framework for natural language analysis has been exploited. GATE is a solution flexible enough to allow the tuning of resources regardless of the NLP logic and it is capable of providing consistent analysis relying on external resources called gazetteers. The Datamuse API are used as an interface to generate such resources, querying semantic repositories such as WordNet, Google Books N-grams and Onelook dictionaries respect to the semantic relations mentioned in Section 2.2.

*3.1.1. Use-case data resources*

Since the application domain for a use-case is a finite set of elements, real taxonomies have been employed under the assumption that the category and the parent category labels of a taxonomy briefly describe the elements in the domain; such labels are considered as all the possible intents of a user query within a fixed domain. Under this assumption the association between a user natural language query and an actual set of documents described by a category label is shifted to the association of the user query to the correct label of the taxonomy. The taxonomy which has been employed for the use case is the well detailed Yelp taxonomy (1318 categories), which is available at the Yelp website. Further experiments to demonstrate the scalability of the methodology have been conducted on the Microsoft Azure Points Of Interest taxonomy (784 categories), the Louisiana Industry Business Type list (170 categories), the Singapore business list (111 categories) and the Greater Newburyport business taxonomy (255 categories). These taxonomies describe the commercial activities and services domain, expressing hierarchical relations between the service type (*parent*) and the service name (*category*) (e.g.: the *Food* service type is parent of several child categories like *Beverage store*, *Cake Shop*, etc.). The relation between the child categories and the parent categories is the *is-a* relation and the reformulation of a question as a category label can then be directly used as input in a structured context to retrieve the actual data.

*3.1.2. NLP technologies and resources*

The resource generation process is implemented using the GATE framework, exploiting its components to build a procedure which generates several corpora improving or worsening query classification with different impact. Each corpus is composed by a set of gazetteers containing terms which are semantically related to the category labels and that are generated from the above mentioned taxonomies. The produced corpora maintain the hierarchical relation which is present in the original taxonomy through the inclusion of lists labeled as the categories into folders labeled as the parent categories. The taxonomy depth consists in two levels including the most general parent category and the most specific category. The labels of the taxonomy are names of commercial activities, services, public administrations, public transportation lines, etc.; such knowledge resources can be integrated in different forms since they are not tied to the classification architecture.

To provide an evaluation of the impact of different kinds of expansion, the resources have been built in different ways by the means of all the available semantic relations. The endpoint which has been employed for accessing semantic databases such as WordNet, Google Books N-grams and Onelook dictionaries is the Datamuse API mentioned in Section 2.2.3. Such interface allows to specify as parameters one or more words, a *rel_[ID]* constraint representing the semantic relation to perform and an optional topic parameter to specify the domain of the word. It is also possible to access all the semantic relations exploiting the API of each repository. For example, using NLTK [50] to access the Wordnet APIs it is possible to get the antonyms of the word *"good"* as: {*evil, ill, evilness, badness, bad*}; the gain in using the Datamuse API lies in the possibility of accessing more relations using one single endpoint and in the presence of a score associated to each retrieved word.

*3.2. Generation of the evaluation corpus*

To give evidence of the classification capabilities when using different semantic expansion criteria a set of lists has been generated, each containing a category label (class) as file name and content. Each file is included in a folder named as its super-category, preserving the taxonomy hierarchy. This set of lists is considered as a reference for the evaluation, behaving as a pattern matching classifier based on gazetteers built without semantic expansion; a similar strategy has been attempted by Raza et al. [51]. To evaluate the impact of each semantic relation in user-intent classification the reference classifier's resources have been expanded using the Datamuse API endpoint. The expansion has been performed by the means of WordNet *synonymy*, *antonymy*, *hypernymy*, *hyponymy*, *holonomy* and *meronymy* relations and the Google Books Ngrams *frequent words associations* (Tables 1, 2). A fully automated procedure to extract the related words by the means of each semantic relation has been produced (its architecture is shown in Fig. 4) and the generated corpus for the input taxonomy is available for research purpose at this link.

Table 1
WordNet semantic relations.

| Relation ID | Relation name | Relation meaning | Example |
|---|---|---|---|
| SYN | Synonymy | *A* **is the same as** *B* | ocean → sea |
| ANT | Antonymy | *A* **is the opposite of** *B* | late → early |
| SPC | Hypernymy | *A* **is kind of** *B* | gondola → boat |
| GEN | Hyponymy | *A* **is more general than** *B* | boat → gondola |
| COM | Holonomy | *A* **comprises** *B* | car → accelerator |
| PAR | Meronymy | *A* **is part of** *B* | trunk → tree |

*3.2.1. Software implementation*

The generation algorithm is composed by four main phases which are: `generateBaseLine`, `generateListsBySemRel`, `enhanceLists` and `getLeavesToSort`.

1. The goal of the first phase is the generation of a reference classifier loaded with a set of files corresponding to the taxonomy categories as resources organized in folders corresponding to the taxonomy parent categories. The files have been generated containing the category labels without any semantic expansion, filtering not-relevant parts of speech like prepositions, conjunctions, articles and punctuation and trimmed by the words which are shorter than 2. The remaining words have been grouped and weighted by the number of occurrences. The aim of this phase is creating a corpus which makes the classification system behave as a pattern matching system once loaded in the GATE pipeline, serving as a reference for evaluating the improvements of each expansion against no expansion.

2. The `generateListsBySemRel` phase makes use of the Datamuse API to get the related words by the means of the relations described in Section 2 for each cluster label. Each label of the taxonomy is expanded exploiting the WordNet relations of *synonymy*, *antonymy*, *hypernymy*, *hyponymy*, *holonomy* and *meronymy* generating a corpus for each semantic relation. Further expansions are provided querying the Onelook dictionaries and the Google Books Ngrams by the means of the *popular nouns with input adjective*, *popular adjectives with input noun*, *popular associated words*, *frequent follower word* and *frequent predecessor word* relations.

3. The `enhanceLists` phase is the most complex, and it is aimed at shaping the content of the files generated in the previous phase meeting the requisites of a gazetteer-based systems and enhancing the classification capabilities through data augmentation. Stop words such as prepositions, conjunctions, articles and punctuation are pruned through POS tagging and further filtering; the words contained in each list are provided with the original lemma, including the parent category labels and the category labels computed in the first phase. Finally, a weight is assigned to the words which have not been computed in the first phase by assigning them a measure of their semantic relatedness respect to the class label. Such measure is expressed by the frequency of the term within a gazetteer after the expansion and the weights are assigned grouping the same words into one and increasing their weight for each repetition.

The above described procedure is implemented by six subphases:

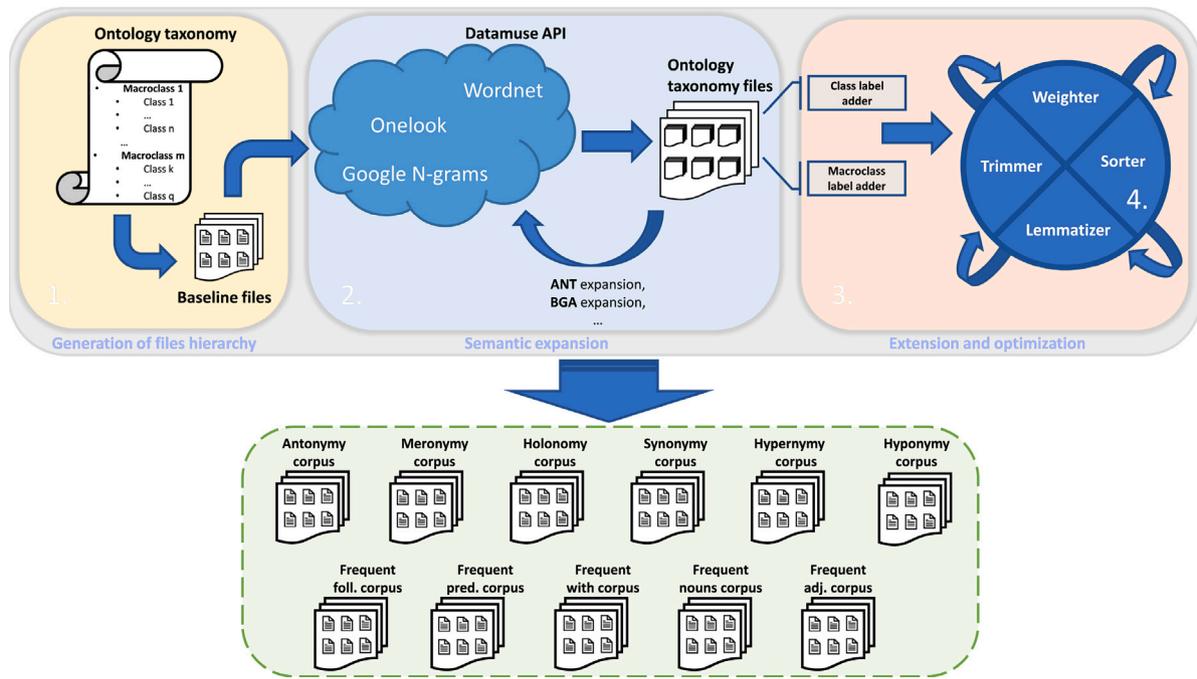

**Fig. 4.** Generation of the corpus.

**Table 2**
Google Books N-grams relations.

| Relation ID | Relation name | Relation meaning | Example |
| --- | --- | --- | --- |
| JJA | Popular nouns | Popular nouns modified by the given adjective | *gradual → increase* |
| JJB | Popular adjectives | Popular adjectives used to modify the given noun | *beach → sandy* |
| TRG | Triggers | Words that are statistically associated with the query word in the same piece of text | *cow → milking* |
| BGA | Frequent followers | $w'$ such that $P(w'|w) \geq 0.001$ | *wreak → havoc* |
| BGB | Frequent predecessors | $w'$ such that $P(w|w') \geq 0.001$ | *havoc → wreak* |

(a) the words related to the category labels (phase 2) are included;
(b) the name of the parent category is added;
(c) a part-of-speech filtering is obtained through the TreeTagger POS-tagger;
(d) the words which are shorter than 2 characters are removed;
(e) the lemmas of the terms are included;
(f) the duplicated words are grouped and followed by the number of occurrences as weight.

As a result of this phase 11 folders are produced, one for each semantic relation described in Section 2; each cluster maintains the hierarchical structure in the form of folders and files which names correspond to the parent category and category labels of the employed taxonomies. Each file contains its label and folder as text, followed by the enhanced output of the semantic repositories obtained through the Datamuse endpoint. It is to be noticed that some of the features included in this phase are also applied to the reference classifier's resources to provide coherence with the rest of the evaluation (e.g. removing the words which are shorter than 2 and making each grouped word followed by its number of occurrences).

4. The last phase aims at providing the most optimized performance in browsing the files, post-processing the lists and sorting them in alphabetical order to be retrieved through a binary search implementation.

The Datamuse API requests are formed using as parameters the *relation ID*, the *category label* of the taxonomy and the *parent category* to provide a more specific field for the expansion.

The procedure is highly scalable, and it can be easily applied to any domain (e.g. using ontology classes or menu items as input taxonomy). Loading the generated resources in a simple framework like the one which has been described in Section 4 allows retrieving any kind of data by natural language knowing only its structure in the form of a taxonomy. The same procedure can be applied to the words which are contained in the actual documents, further extending the retrieval capabilities.

### 3.3. Combination of the generated resources

To provide evidence of the best performing combination of the considered semantic relations for classifying a question as a taxonomy label the resources described in Section 3.2 have been combined in different ways. The combination criteria are all the possible combinations without repetition, providing $\binom{n}{k}$ different configurations with $n = 11$, which is the number of semantic relations that are present in the corpus generated in Section 3.2, and $k$ which is the number of combining relations. The number $n = 11$ is chosen based on the availability of different semantic relations in the leading semantic repositories which are *WordNet*, *Onelook dictionaries* and *Google Books N-grams*. Other semantic relations are available through the Datamuse endpoint mentioned in Section 2.2.3, which are *rhymes*, *approximate rhymes*, *homophones* and *consonant match*; these relations have not been included since their primarily phonetic nature adds little information to words semantics.

To assess the effects on retrieval quality when considering or excluding each relation and its combinations the number of combining relations has been varied from $k = 1$ to $k = n$. The first case for $k = 1$ is represented by the lists obtained in Section 3.2: the number of relations is 11 and there is one corpus for each relation, representing an expansion by the means of a single relation.

$$\binom{n}{k} = \binom{11}{1} = 11 \ combinations$$

For the case $k = 2$ the number of combinations is:

$$\binom{n}{k} = \binom{11}{2} = 55 \ combinations$$

and the possible combinations of the semantic relations (according to the abbreviations shown in Tables 1 and 2) are:

```
                                            SYN TRG
                                       SPC SYN, SPC TRG
                                  PAR SPC, PAR SYN, PAR TRG
                             JJB PAR, JJB SPC, JJB SYN, JJB TRG
                        JJA JJB, JJA PAR, JJA SPC, JJA SYN, JJA TRG
                   GEN JJA, GEN JJB, GEN PAR, GEN SPC, GEN SYN, GEN TRG
              COM GEN, COM JJA, COM JJB, COM PAR, COM SPC, COM SYN, COM TRG
         BGB COM, BGB GEN, BGB JJA, BGB JJB, BGB PAR, BGB SPC, BGB SYN, BGB TRG
    BGA BGB, BGA COM, BGA GEN, BGA JJA, BGA JJB, BGA PAR, BGA SPC, BGA SYN, BGA TRG
ANT BGA, ANT BGB, ANT COM, ANT GEN, ANT JJA, ANT JJB, ANT PAR, ANT SPC, ANT SYN, ANT TRG
```

The cases $k = 3, \dots, k = 11$ follow the same structure and generate $\binom{11}{3} = 165$, $\binom{11}{4} = 330$, $\binom{11}{5} = 462$, $\binom{11}{6} = 462$, $\binom{11}{7} = 330$, $\binom{11}{8} = 165$, $\binom{11}{9} = 55$, $\binom{11}{10} = 11$ and $\binom{11}{11} = 1$ combinations. The latter set is the more complex, describing the combination:

```
ANT, BGA, BGB, COM, GEN, JJA, JJB, PAR, SPC, SYN, TRG
```

that represents the combination of all the considered semantic expansions.

To generate the lists for all the possible combinations of the semantic relations obtained in Section 3.2 a fully automated procedure has been produced and made run in parallel to keep the execution performances attainable.

The lists produced in Section 3.2, representing the categories labels expanded through the 11 semantic relations available on *WordNet*, *Onelook dictionaries* and *Google Books N-grams*, are loaded and kept separated by semantic relation name. Each list of the target corpus is built combining the already obtained lists in $k$ ways: for each $k$, all the lists corresponding to the categories labels are expanded combining their contents by different semantic relations, giving as result a folder named as the considered relations (e.g. ANT, ANT SYN, etc.). Each folder contains the same hierarchy and number of files as the original folders, while the documents within the folder contain the combination of the original lists. To avoid category names replication within the documents, the labels of the categories have been excluded from the computation and added at the end of the process. After the combination process the files are processed by the `enhanceLists` and `getLeavesToSort` libraries described in Section 3.3 to gain coherence with the already generated resources and the assessment which is proposed in Section 4.2.

As a result of the procedure 11 folders are generated, each corresponding to a configuration of $k$ from 1 to 11. Each folder contains the set of all the possible combinations of 11 semantic relations in $k$ ways as directories, each of which preserves the original cluster hierarchy maintaining the parent category labels as folder names and the category labels as file names. The procedure applies a total of 2047 combinations of 11 semantic relations, one for each extraction of $k$ elements, to the input taxonomies. Each combination corresponds to a different tuning for a classifier. The generated resources reach a total of 2 697 946 lists for the Yelp taxonomy and 1 604 848 for the Microsoft Azure Points Of Interest taxonomy. The smaller taxonomies produce 347 990 gazetteers for the Louisiana Industry Business Type list, 227 217 for the Singapore business list and 521 985 for the Greater Newburyport business taxonomy. The generation and the structure of the above described resources is depicted in Fig. 5.

## 4. Evaluation

The assessment is realized making use of the GATE pipeline which is used to classify a set of user questions processed with different corpora loaded as gazetteers. The pipeline is employed as a pseudo-relevance search engine which allows getting from the user questions the information needed to rank them as taxonomy labels. The aim of the assessment is to evaluate the impact of each semantic relation in classifying the questions and to highlight the strengths and weaknesses of different combinations of them. The building of the ground truth is described in Section 4.1. The evaluation tools are described in Section 4.2 and the criteria of the assessment are discussed in Section 4.3. The performance of the reference classifier and the evaluation of the impact of each expansion are proposed in Section 4.4, where the baseline is also compared with other evaluations made through the combination of resources.

### 4.1. Test queries and ground truth

The queries which have been evaluated are a portion of the AOL Query Logs [52] and the ORCAS dataset [53], which contain respectively more than 36 million questions made by users on the popular search engine and 18 million questions derived from the MSMARCO dataset [54]. The sets have been shuffled and reduced to 5000 unique queries, filtering the questions in the form: *"Where can I…?"*, *"Where to find…?"* and queries containing special words like *"…buy…"*, *"…purchase…"* and *"…near…"* which are frequently associated to transactional queries (connected to commercial activities) [55]. This operation has been performed to make the test queries domain the same as the use-case taxonomy domain, describing the field of POIs and local businesses. The number of the filtered queries is a small portion of the complete set, which is coherent with the experiments made by Aloteibi and Sanderson [56] on the Excite search engine, showing that 20% of the users who use search engines have geographical-related intent and the portion of transactional queries among these is even lower. The ground truth has been obtained as an annotation of the test queries with the corresponding taxonomy labels. This operation has been performed through a labeling infrastructure specifically designed for the annotation of the queries (Fig. 7). Such infrastructure requires a set of questions and a set of categories/hyper-categories which are the elements of a taxonomy. Each question is presented to the user, to which is requested to choose an answer from a drop-down menu containing the list of all the hyper-categories; this is a first step that is used to filter the subsequent list of categories from which the user has to choose, also presented as a drop-down menu. In the set of the possible answers are included the *None* and the *Not-an-answer* elements to manage the true negatives and the false negatives in the evaluation phase. The ground truth builder has been used by 50 users annotating 100 queries each with the purpose of associating the intents of the evaluated queries choosing from a set of pre-determined answers. The possible answers correspond to the categories of the taxonomy under evaluation, which in this case is the Yelp taxonomy.

### 4.2. Evaluation infrastructure

The evaluation has been assessed processing the query terms with the GATE engine. The default ANNIE pipeline has been employed, which is composed by the ANNIE tokenizer, the RegEx sentence splitter, the TreeTagger POS-tagger and the ANNIE NE JAPE transducer; the corpora which are described in Section 3.2 and in Section 3.3 have been loaded one at a time as gazetteers (Fig. 6). This approach combines term filtering based on part-of-speech tags and cluster labels expansion based on offline building of gazetteers, which is independent from the classification engine. Such approach is flexible enough to allow the comparison of different strategies for building expansion corpora.

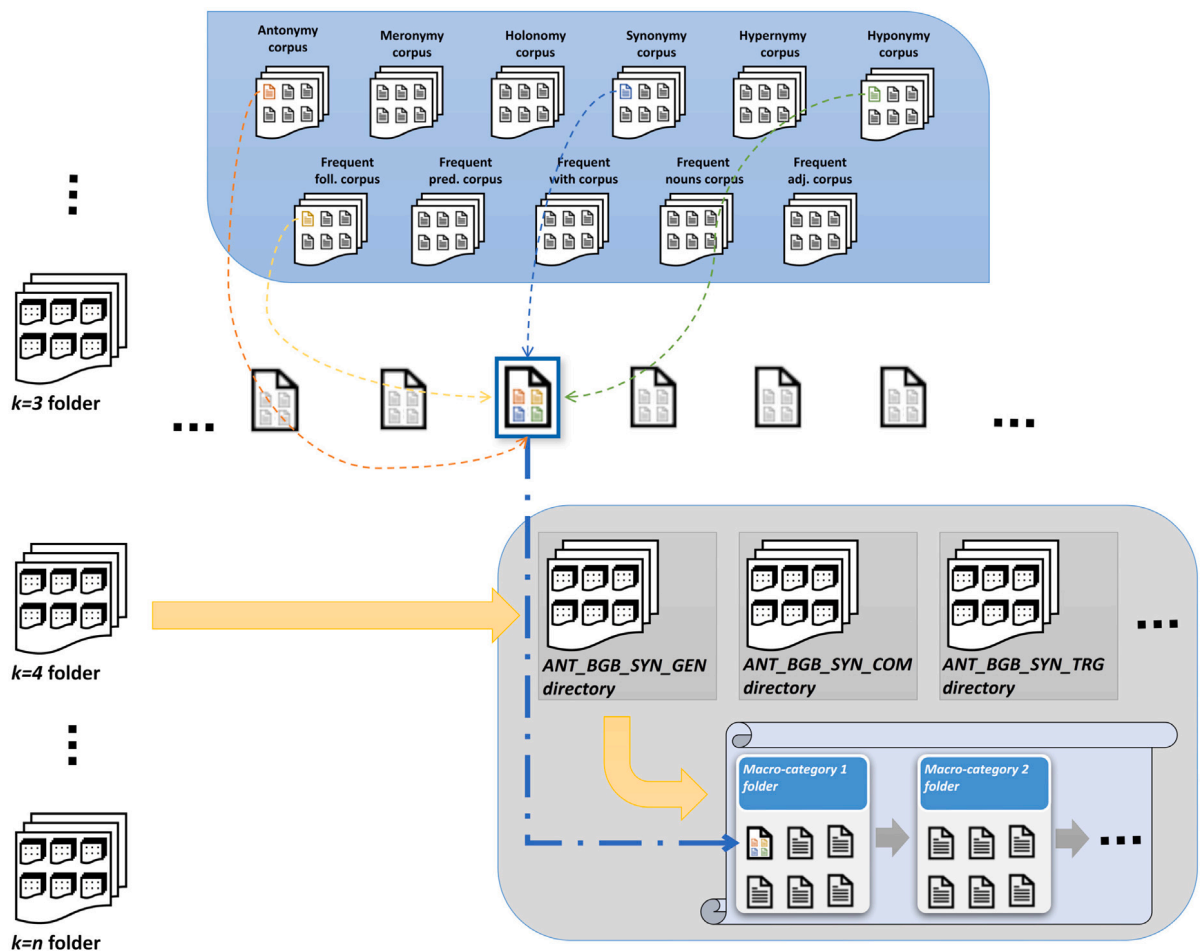

**Fig. 5.** Generation and structure of the combined corpora.

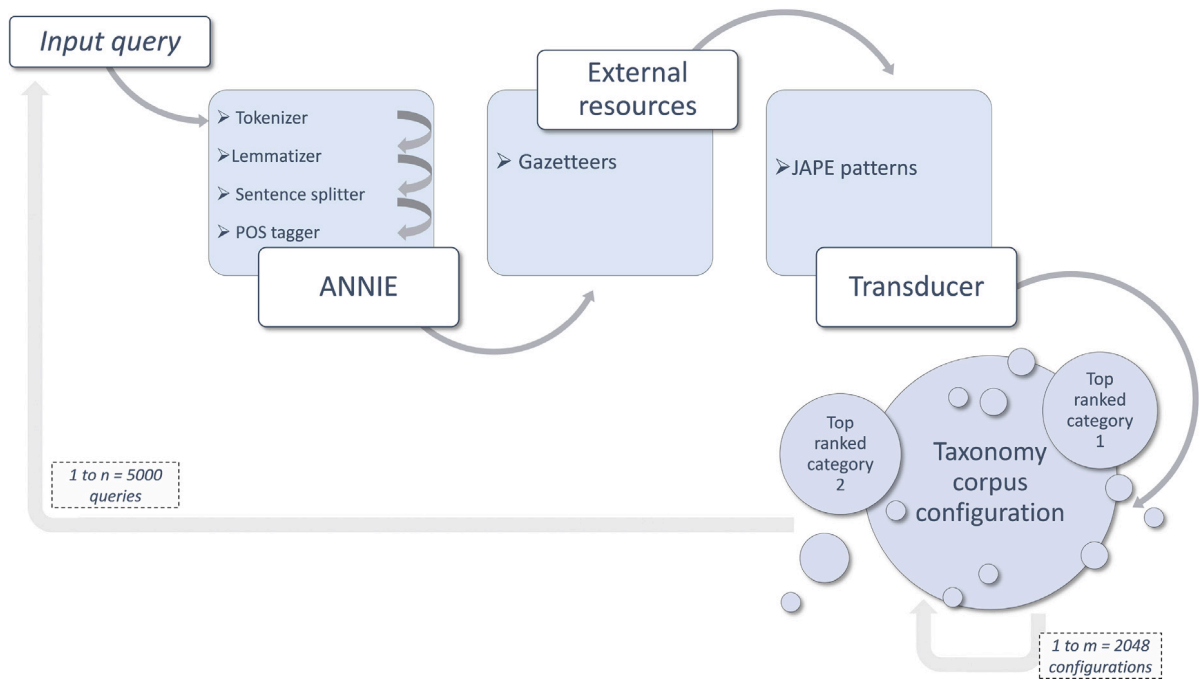

**Fig. 6.** Default ANNIE pipeline loaded with different configurations of corpora.

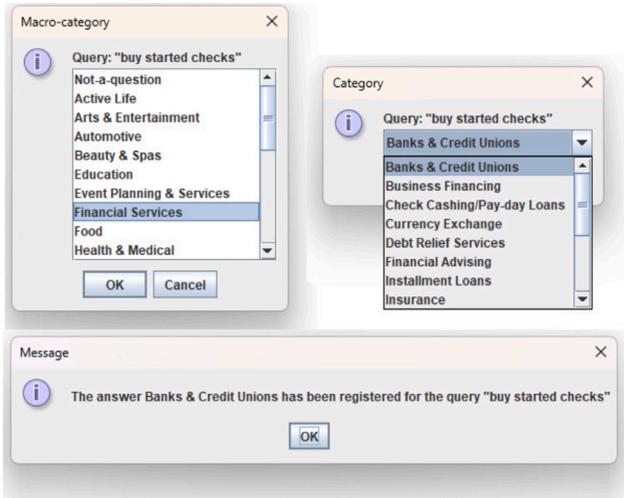

Fig. 7. Ground truth builder infrastructure.

**Table 3**
Reference classifier evaluation.

| Relation | Precision | Recall | F-measure | Accuracy |
|---|---|---|---|---|
| Baseline | 0.62 | 0.29 | 0.40 | 0.25 |

**Table 4**
Performance variation loading one-relation expansions.

| Relation | Precision | Recall | F-measure | Accuracy |
|---|---|---|---|---|
| ANT | −1,95% | −0,21% | −0,38% | −0,21% |
| BGA | −14,07% | +9,87% | +5,12% | +2,84% |
| BGB | −22,66% | +15,80% | +5,42% | +2,86% |
| COM | −3,40% | +0,12% | −0,14% | −0,74% |
| GEN | −6,98% | +2,24% | +0,52% | +1,13% |
| JJA | −12,12% | +8,38% | +4,78% | +1,98% |
| JJB | −15,76% | +6,31% | +1,04% | +0,84% |
| PAR | −1,58% | −0,56% | −0,12% | −0,15% |
| SPC | −11,34% | +4,48% | +0,65% | +0,52% |
| SYN | **+0,32%** | +10,89% | +9,49% | +8,63% |
| TRG | −16,27% | **+36,70%** | **+13,56%** | **+12,50%** |

The assessment is focused on query reformulation and consists in three phases:

1. **Cleaning the query.** The query is preprocessed pruning stop words, conjunctions, articles etc. and lemmatizing words [57] through the GATE framework;
2. **Browsing the corpus.** The extracted query terms are used as search terms browsing the loaded configuration of the corpus. The occurrence of a query term within a list increases the relevance score of the list respect to the user query. If a term is present into a list by multiple semantic relations, this reflects in more significance given to the term respect to the taxonomy label. The sum of the weights constitutes the score of the list and is compared to the rank of the other lists;
3. **Query reformulation.** The query is reformulated as the taxonomy label with the highest score obtained in the second phase respect to the query terms processed in the first phase; a similar strategy can be found in [58]. It is to be noticed that using the entities of an ontology as target taxonomy the elected labels can be directly used to query the underlying structure and retrieve the actual data from the knowledge base.

Each corpus described in Section 3.3 is separately loaded in the pipeline as a set of lists (gazetteers) to obtain the evaluation. The taxonomy categories which are included in the corpus described in Section 3.2 are loaded into the pipeline to provide a reference to compare with the expansions. After setting the evaluation for the reference classifier, the folders which have been generated through the `generateListsBySemRel` function (corresponding to one-relation expansions) are included in the pipeline one at a time and evaluated separately. Afterwards, to compare the combination of each semantic relation with the reference classifier and the one-relation expansions, the assessment is replicated for each combination folder described in Section 3.3 (corresponding to any combination of the 11 relations). A main flowchart of the approach is provided in Fig. 8.

### 4.3. Evaluation criteria

The evaluation of the estimated intents is assessed using the classic information retrieval metrics of precision, recall, F-measure and accuracy. In this case the precision metric represents the capability of the pseudo-relevance search engine to retrieve relevant documents, which are the taxonomy categories, respect to the user query. The recall metric is exploited to measure the capability of the pipeline in retrieving all the relevant documents which can be retrieved by the engine. The F-measure metric is the expression of the harmonic mean of precision and recall, while accuracy quantifies the capacity of the classifier to match expected results. The above described metrics are calculated counting the number of True Positives (TP), False Positives (FP), False Negatives (FN) and True Negatives (TN), which are considered as follows:

- A *true positive* is considered when a taxonomy category which is present in the ground truth is also present in the GATE pipeline outcome;
- a *false positive* is considered when none of the elected categories is present within the ground truth;
- a *true negative* is considered when the elected categories set is empty and the ground truth is empty as well;
- a *false negative* is considered when the elected categories set is empty, while the ground truth is not.

$$\text{Precision} = \frac{TP}{TP + FP} \quad \text{Recall} = \frac{TP}{TP + FN}$$

$$\text{Accuracy} = \frac{TP + TN}{TP + TN + FP + FN}$$

$$\text{F-measure} = 2 \cdot \frac{Precision \cdot Recall}{Precision + Recall}$$

To derive the above described measures as an expression of the impact of each semantic relation in term expansion, the code of the assessment infrastructure has been set to loop among the test queries calculating the metrics by counting TPs, FPs, TNs and FNs. The evaluations are shown in the next section.

### 4.4. Assessment and discussion

The metrics are initially calculated loading the reference classifier and its resources (Section 3.2.1) to evaluate the capabilities of a basic classification system. The same operation is done loading the one-relation expansions corpora (Section 3.2) as resources, quantifying the most impacting relation among those which can be employed as semantic knowledge. A third step has been made loading different combinations of the produced corpora as resources (Section 3.3), highlighting the most effective combination of the semantic relations. All evaluations are made through the techniques discussed in Section 4.3 and are presented as variations respect to the reference classifier, similarly to Azad et al. [11] and Fang [59].

**Reference classifier evaluation.** To provide a reference for the classification system the GATE pipeline has been configured for the assessment to behave as a pattern matching engine, which is the mildest way for classifying the queries as taxonomy labels. Actually, this classifier founds a TP only if a category label is contained in the query. The dataset employed for the baseline classifier is built filling each gazetteer with the relevant words and lemmas of the taxonomy labels, giving them unitary weight; also the parent categories words have been included and half weighted. This reflects in a hit of the classifier whenever a word contained in the query is found in a gazetteer name (or its parent category), which is returned as the user intent. All lists used in the evaluation share the same structure and function, added with the terms obtained by different combinations; the queries have been tokenized and lemmatized as for the rest of the evaluation. The numeric results are shown in Table 3 and are depicted as straight lines in Figs. 9~11, expressing a reference for each evaluation discussed below.

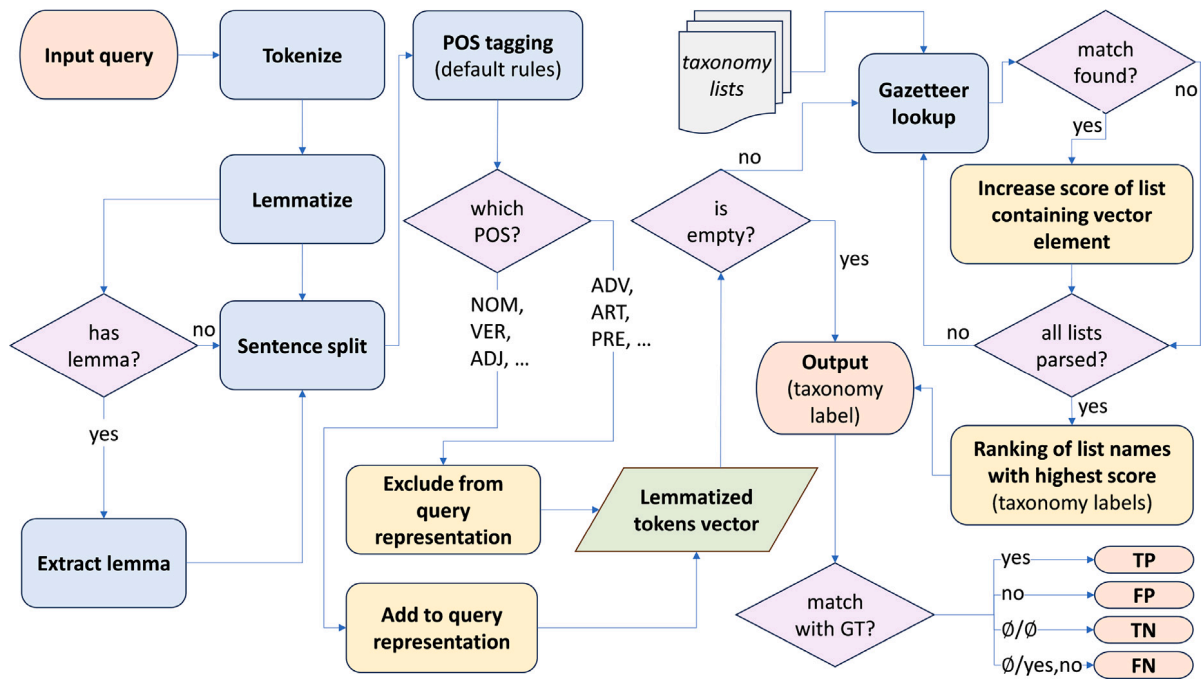

**Fig. 8.** Main flowchart of the approach.

Table 5
Configurations with highest precision wrt the reference classifier.

| Configuration | Variation |
| --- | --- |
| ANT SYN | **+0,35%** |
| ANT PAR SYN | +0,33% |
| SYN | +0,29% |
| PAR SYN | +0,27% |
| ANT COM SYN | −0,02% |
| ANT COM PAR SYN | −0,04% |
| COM SYN | −0,08% |
| COM PAR SYN | −0,09% |
| ANT SPC SYN | −0,33% |
| ANT PAR SPC SYN | −0,35% |

Table 6
Configurations with highest recall wrt the reference classifier.

| Configuration | Variation |
| --- | --- |
| BGA BGB GEN JJA JJB PAR SPC SYN TRG | **+40,79%** |
| ANT BGA BGB GEN JJA JJB PAR SPC SYN TRG | **+40,79%** |
| BGA BGB GEN JJA PAR SPC SYN TRG | +40,74% |
| ANT BGA BGB GEN JJA PAR SPC SYN TRG | +40,74% |
| BGA BGB COM GEN JJA JJB PAR SPC SYN TRG | +40,74% |
| ANT BGA BGB COM GEN JJA JJB PAR SPC SYN TRG | +40,74% |
| BGA BGB COM GEN JJA PAR SPC SYN TRG | +40,70% |
| ANT BGA BGB COM GEN JJA PAR SPC SYN TRG | +40,70% |
| BGA BGB GEN JJA JJB SPC SYN TRG | +40,65% |
| ANT BGA BGB GEN JJA JJB SPC SYN TRG | +40,65% |

Table 7
Configurations with highest accuracy wrt the reference classifier.

| Configuration | Variation |
| --- | --- |
| BGB PAR SYN TRG | **+12.97%** |
| PAR SPC SYN TRG | **+12,97%** |
| ANT BGB PAR SYN TRG | **+12,97%** |
| ANT PAR SPC SYN TRG | **+12,97%** |
| BGB GEN PAR SYN TRG | **+12,97%** |
| BGB PAR SPC SYN TRG | **+12,97%** |
| GEN PAR SPC SYN TRG | **+12,97%** |
| ANT BGB GEN PAR SYN TRG | **+12,97%** |
| ANT BGB PAR SPC SYN TRG | **+12,97%** |
| ANT GEN PAR SPC SYN TRG | **+12,97%** |

Table 8
Configurations with highest F-measure wrt the reference classifier.

| Configuration | Variation |
| --- | --- |
| BGB PAR SYN TRG | **+15.25%** |
| ANT BGB PAR SYN TRG | **+15.25%** |
| BGB GEN PAR SYN TRG | **+15.25%** |
| BGB PAR SPC SYN TRG | **+15.25%** |
| ANT BGB GEN PAR SYN TRG | **+15.25%** |
| ANT BGB PAR SPC SYN TRG | **+15.25%** |
| BGB GEN PAR SPC SYN TRG | **+15.25%** |
| ANT BGB GEN PAR SPC SYN TRG | **+15.25%** |
| PAR SPC SYN TRG | +15,19% |
| ANT PAR SPC SYN TRG | +15,19% |

**One-relation expansion evaluation.** The results are shown in Table 4. This evaluation shows that is held more precision loading a classifier with the synonymy expansion (SYN) of the taxonomy labels. Respect to the reference classifier the precision measure does not hold a significant improvement, probably because pattern matching performs already well in the case of the query logs, including words which are also in the labels of the Yelp taxonomy. The highest recall is reached loading the TRG expansion (words that are statistically associated with the query word in the same piece of text). The highest values for

Table 9
Configurations with lowest precision wrt the reference classifier.

| Configuration | Variation |
|---|---|
| BGA BGB COM GEN JJA JJB SPC TRG | **−23,88%** |
| ANT BGA BGB COM GEN JJA JJB SPC TRG | −23,88% |
| BGA BGB GEN JJA JJB SPC | −23,88% |
| ANT BGA BGB GEN JJA JJB SPC | −23,88% |
| BGA BGB COM GEN JJA JJB SPC | −23,88% |
| ANT BGA BGB COM GEN JJA JJB SPC | −23,88% |
| BGB GEN JJA JJB SPC TRG | −23,88% |
| ANT BGA BGB GEN JJA JJB SPC TRG | −23,88% |
| BGA BGB GEN JJA JJB PAR SPC | −23,88% |
| ANT BGA BGB GEN JJA JJB PAR SPC | −23,88% |

Table 10
Configurations with lowest recall wrt the reference classifier.

| Configuration | Variation |
|---|---|
| ANT | **−0,05%** |
| COM | −0,04% |
| ANT COM | +0,05% |
| PAR | +0,16% |
| ANT PAR | +0,28% |
| COM PAR | +0,30% |
| ANT COM PAR | +0,33% |
| GEN | +2,13% |
| ANT GEN | +2,27% |
| COM GEN | +2,30% |

Table 11
Configurations with lowest accuracy wrt the reference classifier.

| Configuration | Variation |
|---|---|
| COM | **−0,39%** |
| ANT | −0,31% |
| ANT COM | −0,31% |
| PAR | −0,15% |
| COM PAR | −0,15% |
| ANT COM PAR | −0,15% |
| ANT PAR | −0,07% |
| GEN | +0,15% |
| COM GEN | +0,15% |
| ANT COM GEN | +0,15% |

Table 12
Configurations with lowest F-measure wrt the reference classifier.

| Configuration | Variation |
|---|---|
| COM | **−0,59%** |
| ANT | −0,49% |
| ANT COM | −0,49% |
| PAR | −0,27% |
| COM PAR | −0,27% |
| ANT COM PAR | −0,27% |
| ANT PAR | −0,17% |
| GEN | +0,12% |
| COM GEN | +0,12% |
| ANT COM GEN | +0,12% |

F-measure and accuracy are both associated to the TRG expansion, probably because the set of queries is made by real users which follow similar patterns when making questions on specific fields; in particular, the improvement of the F-measure reflects an improvement in retrieving documents.

**Combination of the semantic relations.** The analysis has been extended evaluating all the combinations of the produced corpora as described in Section 3.3: the terms contained in the gazetteers have been combined to produce a set of documents for each combination of $k$ elements, with $k$ ranging from 1 to 11, which is the total number of semantic relations. To provide the analysis the data obtained from the evaluations described in Section 3.3 have been employed.

The evaluation includes the configuration names, the total TP, FP, TN and FN, precision, recall, F-measure and accuracy values. The data have been ordered sorting the configurations by relation name respect to each cardinality subset ($k = 1$, $k = 2$,...) and each configuration label is associated to a numeric id which is also reported in Figs. 9∼11; the vertical sections in such figures represent the size of the different subsets. The purpose of these measurements is to quantify different classification capabilities tuning the classifier with different semantic resources. In Tables 7∼12 are presented the results obtained for the taxonomy under evaluation, showing the most and the less impacting measures found for each configuration respect to the reference classifier. The complete assessment is available for research purposes at this link. The overall best precision enhancement (Table 5) is held by the reference classifier loaded with the ANT SYN combination and more in general by all those configurations including the SYN expansion, which is also the better performing in one-relation analysis (Table 4). All the combinations having high precision measures are associated to the presence of the SYN relation. Considering the combinations in Table 5 and excluding the leading relation SYN, the role of synonyms becomes more noticeable, resulting in a worsening of ∼2.5% for the top combinations and of ∼11% and more for the others. The reason for this can be found in the higher number of elements in the set of synonyms compared to the other relations, due to their higher availability. Furthermore, according to the Datamuse APIs scores described in Section 3.1.2 the distance between terms related by synonymy is lower compared to the one linking terms by other relations; this results in a higher correlation of synonyms to the original concepts and thus in better retrieval capabilities. Some regularities which are highlighted by the Precision plot (Fig. 9) are associated to the ANT, PAR, SPC and SYN relations, which occur in the local peaks of each set.

The improvements in recall (Fig. 10) are significant; this was expected since Sharma et al. [60] demonstrates that the expansion of queries based on term matching improves the recall performance. The highest recall increase is achieved employing the BGA BGB GEN JJA JJB PAR SPC SYN TRG configuration, showing significantly higher values with more relations included; the top 10 results (Table 6) include the most numerous configuration and overcome any result obtained with one-relation expansions. The F-measure evaluations (Fig. 11) show that the best results are obtained by the BGB PAR SYN TRG configuration which is, despite the evident improvement respect to the reference classifier, not far by the other measurements and this is also the case for the accuracy evaluations (Fig. 12) which are also headed by the BGB PAR SYN TRG configuration. F-measure and Accuracy plots express similar behaviors across the sets, showing an increasing trend associated to the PAR, SPC, SYN and TRG relations which is more pronounced on the SYN and TRG relations without which is present a steep decrease in performances. The highest accuracy values (Table 7) show an increase of about 13% respect to the reference classifier, while the highest F-measure values (Table 8) are steady on 15% increase; both increments are associated to the PAR SYN TRG relations. It is to be noticed that considering the results in terms of precision, recall, f-measure and accuracy the registered values are very low since the measurements are tightly bound to the performance of the reference classifier which in this case is simple text match; this is noticeable looking at the values of the reference classifier evaluations. This choice has been made since the target of the study is to evaluate the impact of different semantic relations and not to evaluate the performances of a specific classifier.

The lower performing combinations of resources are reported in Tables 9∼12, in which are highlighted the 10 combinations which perform worse respect to the reference classifier. The 10 lowest precision values are constantly 23,88% lower respect to the baseline. It is to be noticed that none of the worst performing configurations contain the SYN relation, which has been identified as a classification-enhancing relation. In Table 10 most of the recall values are positive, which means that the reference classifier has better recall capabilities respect to any semantic expansion. The negative values are associated to the ANT and COM relations, which denote a higher tendency to miss relevant results. It is to be noticed that the combinations with lower precision are made of larger sets of relations, which indicates that the inclusion of less effective relations deteriorates the classifier's positive predictions. The influence of the number of relations is particularly noticeable looking at Figs. 9 and 10: the recurring patterns through the subsets are well visible and the leftmost sparsity of the precision and recall plots correspond to a reduced number of relations included. The lowest values of accuracy and F-measure shown in Tables 11 and 12 include the same relations, although with different values. The reason is that for the less enhancing relations the expansion lists are mostly empty and where no improvement is applied, the expansion results in a light worsening respect to the reference classifier. A sparse trend of data when employing a low number of relations is noticeable for each measure, corresponding to a high density when using more relations. This is caused by the fact that some expansions have few term contributions or none at all, hence adding the terms of those expansions does not enhance the measures under evaluation. For the same reason the configurations including a high number of relations do not correspond to the highest results of precision and F-measure.

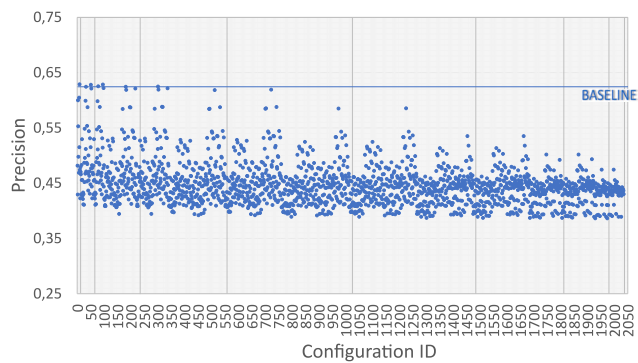

**Fig. 9.** Precision plot for multi-relation expansions.

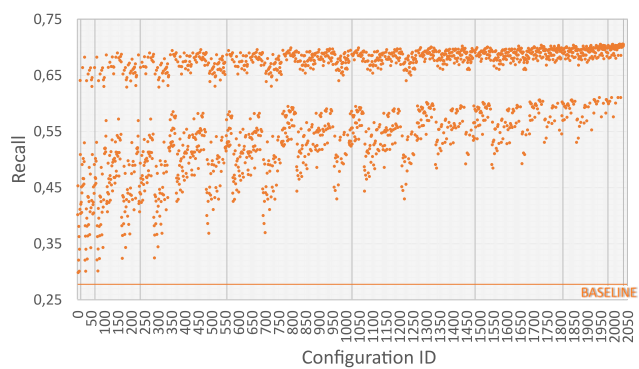

**Fig. 10.** Recall plot for multi-relation expansions.

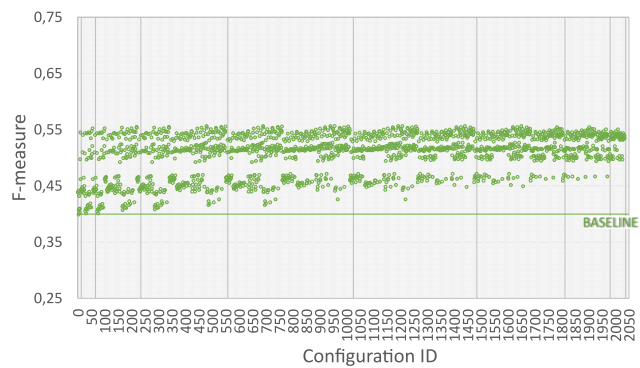

**Fig. 11.** F-measure plot for multi-relation expansions.

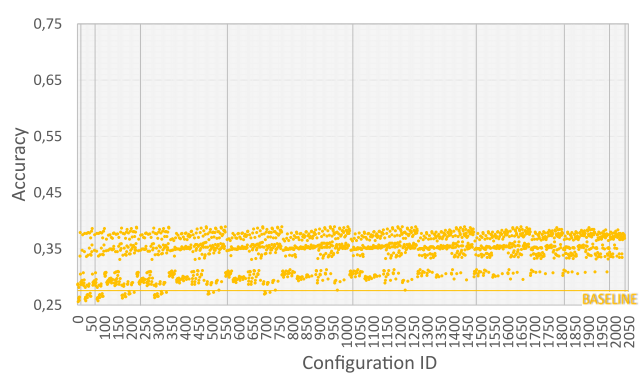

**Fig. 12.** Accuracy plot for multi-relation expansions.

## 5. Conclusions

In this paper the main concepts in the field of automatic term expansion have been discussed; the fields of relevance feedback and search results clustering have been investigated in detail. Automatic term expansion often deals with the latent semantics of the terms and knowledge resources like WordNet, the Onelook dictionaries and the Google Books N-grams databases are used to get the latent senses. The relations which are accessible through such repositories have been exploited using an automatic procedure aimed at semantically expanding the labels of an input taxonomy. The taxonomy structure has been used as a source of information expanding its labels by the means of different semantic relations, then a further step has been made to combine the extracted resources. To obtain a comparison of the effects in employing the different assets, the generated resources have been integrated as gazetteers into several GATE pipelines which are used as pseudo-relevance feedback systems to reformulate a set of queries as taxonomy labels. The evaluation provides numeric evidence of the most effective combination of the semantic relations used as term expansion resources and shows the best trade-off between accuracy and noise introduction. This study highlights the different impact of semantic relations and their combinations when used to classify natural language questions as classes of a hierarchical structure. Based on these results, this study demonstrates which semantic expansion resources are better suited for enhancing the semantic capabilities of search engines. The generation and the combination procedures are completely automated and provide high scalability: the system can be input with any taxonomy and it can be applied to any hierarchical structure like indexes, text structures, ontology taxonomies, etc. The generated resources have been demonstrated to be exploitable to classify any natural language query as labels of an input taxonomy, enhancing the retrieval of any data underneath and using the most effective combination of the available semantic resources.

## CRediT authorship contribution statement

**Lorenzo Massai:** Writing – review & editing, Writing – original draft, Visualization, Validation, Supervision, Software, Resources, Project administration, Methodology, Investigation, Funding acquisition, Formal analysis, Data curation, Conceptualization.

## Declaration of competing interest

The authors declare that they have no known competing financial interests or personal relationships that could have appeared to influence the work reported in this paper.

**Data availability**

All the data employed for the assessment including 5000 unique queries with the ground truth, the files corresponding to each semantic relation containing the expanded taxonomy labels, the complete Precision/Recall/F-measure/Accuracy evaluation for each classifier, the code for collecting the ground truth, for generating the single relations expansions and for combining the relations are referred in the paper and made available at these links: evaluation, single relation expansions and code.

**Acknowledgments**

A special thanks is owed to the Yelp, Microsoft Research and AOL organizations who made available the huge amount of data which are exploited to build the structure of the corpora for the use-case evaluation and the queries which have been used in the assessment. Most of the data which are exploited to compute such resources is obtained through the WordNet framework, made available by the Princeton University which is kindly acknowledged.

**References**

[1] M. Raza, R. Mokhtar, N. Ahmad, A survey of statistical approaches for query expansion, Knowl. Inf. Syst. 61 (2019) http://dx.doi.org/10.1007/s10115-018-1269-8.

[2] D. Sharma, R. Pamula, D. Chauhan, Semantic approaches for query expansion, Evol. Intell. 14 (2021) http://dx.doi.org/10.1007/s12065-020-00554-x.

[3] M. Zingla, C. Latiri, P. Mulhem, C. Berrut, Y. Slimani, Hybrid query expansion model for text and microblog information retrieval, Inf. Retr. J. 21 (2018) http://dx.doi.org/10.1007/s10791-017-9326-6.

[4] L. Lopes, J. Gadge, Hybrid approach for query expansion using query log, Int. J. Appl. Inf. Syst. 7 (6) (2014) 30–35, Published by Foundation of Computer Science, New York, USA.

[5] J. Bhogal, A. MacFarlane, P. Smith, A review of ontology based query expansion, Inf. Process. Manag. 43 (4) (2007) 866–886, URL: http://dblp.uni-trier.de/db/journals/ipm/ipm43.html#BhogalMS07.

[6] A. Sihvonen, P. Vakkari, Subject knowledge improves interactive query expansion assisted by a thesaurus, J. Doc. 60 (2004) 673–690, http://dx.doi.org/10.1108/00220410410568151.

[7] L. Dey, S. Singh, R. Rai, S. Gupta, Ontology aided query expansion for retrieving relevant texts, 2005, pp. 126–132, http://dx.doi.org/10.1007/11495772_21.

[8] T. Hofmann, Unsupervised learning by probabilistic latent semantic analysis, Mach. Learn. 42 (1–2) (2001) 177–196.

[9] E.N. Efthimiadis, Interactive query expansion: A user-based evaluation in a relevance feedback environment, J. Am. Soc. Inf. Sci. 51 (11) (2000) 989–1003, http://dx.doi.org/10.1002/1097-4571(2000)9999:9999%3C::AID-ASI1002%3E3.0.CO;2-B.

[10] R. Kumar, G. Bhanodai, R. Pamula, Book search using social information, user profiles and query expansion with pseudo relevance feedback, Appl. Intell. 49 (2019) http://dx.doi.org/10.1007/s10489-018-1383-z.

[11] H.K. Azad, A. Deepak, C. Chakraborty, K. Abhishek, Improving query expansion using pseudo-relevant web knowledge for information retrieval, Pattern Recognit. Lett. 158 (2022) 148–156, http://dx.doi.org/10.1016/j.patrec.2022.04.013, URL: https://www.sciencedirect.com/science/article/pii/S0167865522001088.

[12] R. Crimp, A. Trotman, Automatic term reweighting for query expansion, 2017, pp. 1–4, http://dx.doi.org/10.1145/3166072.3166074.

[13] Elvina, R. Mandala, Improving effectiveness information retrieval system using pseudo irrelevance feedback, in: 2020 IEEE International Conference on Sustainable Engineering and Creative Computing (ICSECC), 2020, pp. 463–468, http://dx.doi.org/10.1109/ICSECC51444.2020.9557550.

[14] J. Devlin, M. Chang, K. Lee, K. Toutanova, BERT: pre-training of deep bidirectional transformers for language understanding, in: J. Burstein, C. Doran, T. Solorio (Eds.), Proceedings of the 2019 Conference of the North American Chapter of the Association for Computational Linguistics: Human Language Technologies, NAACL-HLT 2019, Minneapolis, MN, USA, June 2-7, 2019, Volume 1 (Long and Short Papers), Association for Computational Linguistics, 2019, pp. 4171–4186, http://dx.doi.org/10.18653/v1/n19-1423.

[15] X. Wang, C. Macdonald, N. Tonellotto, I. Ounis, Pseudo-relevance feedback for multiple representation dense retrieval, in: Proceedings of the 2021 ACM SIGIR International Conference on Theory of Information Retrieval, ICTIR '21, Association for Computing Machinery, New York, NY, USA, 2021, pp. 297–306, http://dx.doi.org/10.1145/3471158.3472250.

[16] J. Wang, M. Pan, T. He, X. Huang, X. Wang, X. Tu, A pseudo-relevance feedback framework combining relevance matching and semantic matching for information retrieval, Inf. Process. Manage. 57 (2020) 102342, http://dx.doi.org/10.1016/j.ipm.2020.102342.

[17] M. Ahmed, R. Seraj, S. Islam, The k-means algorithm: A comprehensive survey and performance evaluation, Electronics 9 (2020) 1295, http://dx.doi.org/10.3390/electronics9081295.

[18] V. Chauhan, K. Dahiya, A. Sharma, Problem formulations and solvers in linear SVM: a review, Artif. Intell. Rev. 52 (2019) 803–855, http://dx.doi.org/10.1007/s10462-018-9614-6.

[19] J. Shen, Z. Shen, C. Xiong, C. Wang, K. Wang, J. Han, TaxoExpan: Self-supervised taxonomy expansion with position-enhanced graph neural network, in: Proceedings of the Web Conference 2020, WWW '20, Association for Computing Machinery, New York, NY, USA, 2020, pp. 486–497, http://dx.doi.org/10.1145/3366423.3380132.

[20] O. Kurland, L. Lee, C. Domshlak, Better than the real thing? Iterative pseudo-query processing using cluster-based language models, 2006, CoRR abs/cs/0601046. arXiv:cs/0601046. URL: http://arxiv.org/abs/cs/0601046.

[21] C. Buckley, Why current IR engines fail, Inf. Retr. 12 (2004) 652–665, http://dx.doi.org/10.1007/s10791-009-9103-2.

[22] R. Navigli, S.P. Ponzetto, BabelNet: The automatic construction, evaluation and application of a wide-coverage multilingual semantic network, Artificial Intelligence 193 (2012) 217–250, http://dx.doi.org/10.1016/j.artint.2012.07.001.

[23] J. Bai, D. Song, P. Bruza, J.-Y. Nie, G. Cao, Query expansion using term relationships in language models for information retrieval, in: O. Herzog, H.-J. Schek, N. Fuhr, A. Chowdhury, W. Teiken (Eds.), CIKM, ACM, 2005, pp. 688–695, URL: http://dblp.uni-trier.de/db/conf/cikm/cikm2005.html#BaiSBNC05.

[24] G. Cao, J.-y. Nie, J. Gao, S. Robertson, Selecting good expansion terms for pseudo-relevance feedback, 2008, pp. 243–250, http://dx.doi.org/10.1145/1390334.1390377.

[25] S. Dahir, A. El Qadi, A query expansion method based on topic modeling and dbpedia features, Int. J. Inf. Manag. Data Insights 1 (2021) 100043, http://dx.doi.org/10.1016/j.jjimei.2021.100043.

[26] L. Guo, X. Su, L. Zhang, G. Huang, X. Gao, Z. Ding, Query expansion based on semantic related network, in: 15th Pacific Rim International Conference on Artificial Intelligence, Nanjing, China, August 28–31, 2018, Proceedings, Part II, 2018, pp. 19–28, http://dx.doi.org/10.1007/978-3-319-97310-4_3.

[27] J. Singh, M. Prasad, o. Prasad, J. Er, A. Saxena, C.-T. Lin, A novel fuzzy logic model for pseudo-relevance feedback-based query expansion, Int. J. Fuzzy Syst. 18 (2016) 980–989, http://dx.doi.org/10.1007/s40815-016-0254-1.

[28] W. Alromima, I. Moawad, R. Elgohary, M. Aref, Ontology-based query expansion for arabic text retrieval, Int. J. Adv. Comput. Sci. Appl. (IJACSA) 7 (2016) 223–230, http://dx.doi.org/10.14569/IJACSA.2016.070830.

[29] C. Fellbaum (Ed.), Wordnet: an electronic lexical database, MIT Press, 1998.

[30] Z. Gong, C. Cheang, R.L.H. Uu, Web query expansion by WordNet, 2005, pp. 166–175, http://dx.doi.org/10.1007/11546924_17.

[31] M. Song, I.-Y. Song, X. Hu, R. Allen, Integration of association rules and ontologies for semantic query expansion, Data Knowl. Eng. 63 (2007) 63–75, http://dx.doi.org/10.1016/j.datak.2006.10.010.

[32] R. Raj, B. Sam, A. Shetty, A hybrid framework to refine queries using ontology, Indian J. Sci. Technol. 8 (2015) http://dx.doi.org/10.17485/ijst/2015/v8i24/80883.

[33] H.K. Azad, A. Deepak, Query expansion techniques for information retrieval: A survey, Inf. Process. Manag. 56 (5) (2019) 1698–1735, URL: http://dblp.uni-trier.de/db/journals/ipm/ipm56.html#AzadD19.

[34] S. Jain, S. K.R., R. Jindal, A fuzzy ontology framework in information retrieval using semantic query expansion, Int. J. Inf. Manag. Data Insights 1 (2021) 100009, http://dx.doi.org/10.1016/j.jjimei.2021.100009.

[35] Z. Shi, B. Gu, F. Popowich, A. Sarkar, Synonym-based expansion and boosting-based re-ranking: A two-phase approach for genomic information retrieval, 2005.

[36] R. Chauhan, R. Goudar, R. Sharma, A. Chauhan, Domain ontology based semantic search for efficient information retrieval through automatic query expansion, 2013, pp. 397–402, http://dx.doi.org/10.1109/ISSP.2013.6526942,

[37] R. Navigli, P. Velardi, An analysis of ontology-based query expansion strategies, in: Workshop on Adaptive Text Extraction and Mining, 2003.

[38] D. Buscaldi, P. Rosso, E. Arnal, A wordnet-based query expansion method for geographical information retrieval, 2005.

[39] M. Esposito, E. Damiano, A. Minutolo, G. De Pietro, H. Fujita, Hybrid query expansion using lexical resources and word embeddings for sentence retrieval in question answering, Inform. Sci. 514 (2019) http://dx.doi.org/10.1016/j.ins.2019.12.002.

[40] M. Franco-Salvador, P. Rosso, R. Navigli, A knowledge-based representation for cross-language document retrieval and categorization, in: 14th Conference of the European Chapter of the Association for Computational Linguistics 2014, EACL 2014, 2014, pp. 414–423, http://dx.doi.org/10.3115/v1/E14-1044.

[41] M. Copeland, J. Soh, A. Puca, M. Manning, D. Gollob, Microsoft Azure, Apress, New York, NY, USA, 2015, pp. 3–26.

[42] H. Azad, A. Deepak, A new approach for query expansion using wikipedia and WordNet, Inform. Sci. 492 (2019) http://dx.doi.org/10.1016/j.ins.2019.04.019.

[43] D. Pal, M. Mitra, K. Datta, Improving query expansion using WordNet, J. Assoc. Inf. Sci. Technol. 65 (2014) http://dx.doi.org/10.1002/asi.23143.

[44] Datamuse, Resource available online, 2016, URL: http://www.datamuse.com/api/.

[45] K. Clink, Onelook: Dictionary search, Ref. Rev. 21 (8) (2007) 35, http://dx.doi.org/10.1108/09504120710838912.

[46] H. Cunningham, D. Maynard, K. Bontcheva, V. Tablan, GATE: A Framework and Graphical Development Environment for Robust NLP Tools and Applications, in: Proceedings of the 40th Anniversary Meeting of the Association for Computational Linguistics (ACL'02), 2002.

[47] H. Schmid, Probabilistic part-of-speech tagging using decision trees, in: Proceedings of the International Conference on New Methods in Language Processing, Manchester, UK, 1994.

[48] M. Porter, Snowball: A language for stemming algorithms, 2001, Retrieved March 1.

[49] D. Chen, C. Manning, A fast and accurate dependency parser using neural networks, EMNLP (2014) 740–750.

[50] S. Bird, E. Klein, E. Loper, Natural Language Processing with Python: Analyzing Text with the Natural Language Toolkit, O'Reilly, Beijing, 2009, URL: http://www.nltk.org/book. doi:http://my.safaribooksonline.com/9780596516499.

[51] M. Raza, R. Mokhtar, N. Ahmad, M. Pasha, U. Pasha, A taxonomy and survey of semantic approaches for query expansion, IEEE Access PP (2019) 1, http://dx.doi.org/10.1109/ACCESS.2019.2894679.

[52] G. Pass, A. Chowdhury, C. Torgeson, A picture of search, in: X. Jia (Ed.), Proceedings of the 1st International Conference on Scalable Information Systems, Infoscale 2006, Hong Kong, May 30-June 1006, in: ACM International Conference Proceeding Series, vol. 152, ACM, 2006, p. 1, http://dx.doi.org/10.1145/1146847.1146848.

[53] N. Craswell, D. Campos, B. Mitra, E. Yilmaz, B. Billerbeck, ORCAS: 18 million clicked query-document pairs for analyzing search, 2020, CoRR abs/2006.05324. arXiv:2006.05324. URL: https://arxiv.org/abs/2006.05324.

[54] T. Nguyen, M. Rosenberg, X. Song, J. Gao, S. Tiwary, R. Majumder, L. Deng, MS MARCO: a human generated machine reading comprehension dataset, 2016, CoRR abs/1611.09268. arXiv:1611.09268. URL: http://arxiv.org/abs/1611.09268.

[55] B.J. Jansen, D.L. Booth, A. Spink, Determining the informational, navigational, and transactional intent of web queries, Inf. Process. Manag. 44 (3) (2008) 1251–1266, http://dx.doi.org/10.1016/j.ipm.2007.07.015.

[56] S. Aloteibi, M. Sanderson, Analyzing geographic query reformulation: An exploratory study, J. Assoc. Inf. Sci. Technol. 65 (2014) http://dx.doi.org/10.1002/asi.22961.

[57] S. Larson, A. Mahendran, J. Peper, C. Clarke, A. Lee, P. Hill, J. Kummerfeld, K. Leach, M. Laurenzano, L. Tang, J. Mars, An evaluation dataset for intent classification and out-of-scope prediction, 2019, pp. 1311–1316, http://dx.doi.org/10.18653/v1/D19-1131.

[58] R. Kumar, S. Sharma, Smart information retrieval using query transformation based on ontology and semantic-association, Int. J. Adv. Comput. Sci. Appl. 13 (2022) http://dx.doi.org/10.14569/IJACSA.2022.0130446.

[59] H. Fang, A re-examination of query expansion using lexical resources, 2008, pp. 139–147.

[60] D. Sharma, R. Pamula, D. Chauhan, A contemporary combined approach for query expansion, Multimedia Tools Appl. 81 (2020) http://dx.doi.org/10.1007/s11042-020-09172-2.